%% file: PaperForReview.tex
\documentclass[10pt,twocolumn,letterpaper]{article}

\usepackage[pagenumbers]{cvpr} % To force page numbers, e.g. for an arXiv version

\usepackage{graphicx}
\usepackage{amsmath}
\usepackage{amssymb}
\usepackage{booktabs}
\usepackage{xcolor}
\usepackage{multirow}
\usepackage{makecell}
\usepackage{url}
\usepackage{subcaption}
\usepackage{bm}
\usepackage[pagebackref,breaklinks,colorlinks]{hyperref}

\usepackage[capitalize]{cleveref}
\crefname{section}{Sec.}{Secs.}
\Crefname{section}{Section}{Sections}
\Crefname{table}{Table}{Tables}
\crefname{table}{Tab.}{Tabs.}

\def\cvprPaperID{5285} % *** Enter the CVPR Paper ID here
\def\confName{CVPR}
\def\confYear{2022}
\newcommand{\xt}[1]{\textcolor{black}{#1}}
\newcommand{\bp}[1]{\textcolor{black}{#1}}
\newcommand{\deli}[1]{\textcolor{black}{#1}}

\begin{document}

%\title{Efficient Mixed Attention For Vision Transformer}
% \title{Rethinking Dynamic Attention in Vision Transformer}
%\title{Static Priors of Dynamic Attention in Vision Transformer}
%\title{Efficient Vision Transformer with Attention Distillation and Sparsification}
\title{Accelerating Vision Transformers Based on  Heterogeneous Attention Patterns}

% \and

\author{
Deli Yu\thanks{Equal contribution.}\quad Teng Xi$^{*}$\quad  Jianwei Li$^{*}$ \quad Baopu Li\thanks{Corresponding author.}
\quad Gang Zhang \\ 
Haocheng Feng \quad Junyu Han \quad Jingtuo Liu \quad Errui Ding \quad Jingdong Wang\\
Department of Computer Vision Technology(VIS), Baidu Inc.\\
{\tt\small $\{$yudeli, xiteng01, lijianwei04, baopuli, zhanggang03$\}$@baidu.com} \\
{\tt\small $\{$fenghaocheng, hanjunyu, liujingtuo, dingerrui, wangjingdong$\}$@baidu.com}
}

\maketitle
%%%%%%%%% ABSTRACT
\begin{abstract}
\bp{Recently, Vision Transformers~(ViTs) have attracted a lot of attention in the field of computer vision.}
Generally, the powerful representative capacity of ViTs mainly benefits from the self-attention mechanism, which has a high computation complexity.
To accelerate ViTs, we propose an integrated compression pipeline 
based on observed heterogeneous attention  patterns across layers. 
On one hand, different images share more similar attention patterns in  early layers than  later layers, indicating that the dynamic query-by-key self-attention matrix may be replaced with a static self-attention matrix  in early layers. 
Then, we propose a dynamic-guided static self-attention~(DGSSA) method where the  matrix inherits self-attention information from the replaced dynamic self-attention 
\bp{to effectively improve the feature representation ability of ViTs. }
On the other hand,  the attention maps have more low-rank patterns, which reflect token redundancy, in later layers than early layers. In a view of linear dimension reduction, we \bp{further} propose a method of global aggregation pyramid~(GLAD) to reduce the number of tokens in later layers of ViTs,  such as  Deit. Experimentally, the integrated  compression pipeline of DGSSA and GLAD can accelerate up to 121\% run-time throughput compared with DeiT, which surpasses all SOTA approaches. %and 10\% run-time throughput of Deit-Small and Swin-Small, respectively, with negligible accuracy drop\xt{\deli{, }which significantly surpass\deli{es} all state-of-the-art approaches.}

\end{abstract}

%%%%%%%%% BODY TEXT
\section{Introduction}
\label{sec:intro}

\bp{Recently, Vision Transformers~(ViTs) have shown their impressive capabilities in many computer vision tasks such as image classification ~\cite{dosovitskiy2020image,touvron2021training,liu2021swin,jiang2021token, yuan2021tokens, srinivas2021bottleneck, chen2021pre, han2021transformer, chu2021conditional, dosovitskiy2020image}, object detection~\cite{carion2020end_detr,dai2021dynamic, zhu2020deformable, dai2021up, carion2020end},  semantic segmentation~\cite{xie2021segformer,YuanFHLZCW21,zheng2021rethinking, wang2021end} and so on.}
% Vision Transformers~(ViTs) has shown priority over Convolutional Neural Networks~(CNNs) in image classification task of  ILSVRC-2012. 
% % when meeting large scale data in Computer Vision. 
% % ViTs even can defeat CNNs with elaborated data augmentations and training settings in image classification task of  ILSVRC-2012. 
% Moreover, ViTs also produces remarkable performance in other visual recognition task, such object detection, semantic segmentation, etc. 
The impressive representative capacity of ViTs mainly comes from the self-attention~(SA) mechanism 
% where the attention map is calculated to attend
\deli{of capturing }
\bp{the long range semantic}
\deli{dependencies, }
% region for each image, 
\bp{which are difficult for previous convolution neural networks~(CNNs).}
% the main visual semantic region for each image in a dynamic manner.
\xt{However, the quadratic calculation complexity of self-attention still limits its further applications}, \bp{especially on resource-constrained scenarios.} 

% Obviously, the intrinsic characteristic of attention patterns \bp{may be of favor} to 
% % designing more efficient and stronger ViTs. 
% \deli{make ViTs more efficient and stronger. }
% \bp{Despite recent great advantages in different transformer-related architecture design such as Swin-Transformer~\cite{liu2021swin}, CvT~\cite{wu2021cvt}, ConViT~\cite{d2021convit}, MobileViT~\cite{mehta2021mobilevit}, Mobile-former~\cite{chen2021mobile}, there still lacks an in-depth understanding or analysis about the attention patterns for 
% % transformers in the field of computer vision.
% \deli{ViTs. }
% }

% Thus, there is an open question  accelerate the calculation of attention map.  
% attention patterns of Transformers in NLP tasks. 

% \bp{Specifically, inspired by some recent works~\cite{raganato2020fixed,tay2021synthesizer} in the field of NLP, we argue that the attention mechanism can be explained in a better manner from the perspective of \deli{mixture of} dynamic and static attention.} The process of attention is conducted in a dynamic manner \bp{that} means the calculated attention matrix of different images are different due to the content-based style of self-attention. 
%  By contrast, 
% % the content-based dynamic attention matrix is replaced with 
% \bp{？a content-free  matrix is shared by all images in the regime of static attention.}

To achieve efficient computation, 
the  standard self-attention is modified with a new form of sparse   versions~\cite{parmar2018image,child2019generating_sparse_transformer,beltagy2020longformer, qiu2019blockwise, kitaev2020reformer, ainslie2020etc, zaheer2020big}. 
% or low-rank self-attention~\cite{wang2020linformer, zhu2021long}.
% In addition to sparse form, 
Meanwhile, 
some \bp{efforts}~\cite{tolstikhin2021mlp,raganato2020fixed} rethink the necessity of dynamic self-attention and replace it with a form of \deli{static self-attention~(SSA) matrix}, which is shared by all images. Thus, the repeated computation of dynamic self-attention matrix can be saved.  Static form is more  hardware-friendly than sparse form in run-time mode, due to no
dependencies on specialized hard-and soft-ware framework. 
Since the complexity of self-attention is related to token numbers, 
\deli{another line  prunes~\cite{tang2021patch,rao2021dynamicvit,xu2021evo} or aggregaes~\cite{wang2021pyramid,liu2021swin,wu2021cvt,chu2021twins} tokens, and can get a slimmed pyramid distribution of token numbers and largely increase run-time throughput too.}
%To be specific, \deli{MLP-Mixer~\cite{tolstikhin2021mlp} and Synthesizer~\cite{tay2021synthesizer} 
%use multilayer perceptrons~(MLPs) and synthetic random matrix, respectively.}
%Thus, repeated calculation of dynamic self-attention can be saved a lot.
%  it may be reasonable to replace dynamic attention with a shared static one without much degradation of model capacity in early layers \bp{since such an operation may save the computations of the attention map a lot.} 

 \begin{figure}[t]
  \centering
%   \fbox{\rule{0pt}{2in} \rule{0.9\linewidth}{0pt}}
  \includegraphics[width=\linewidth]{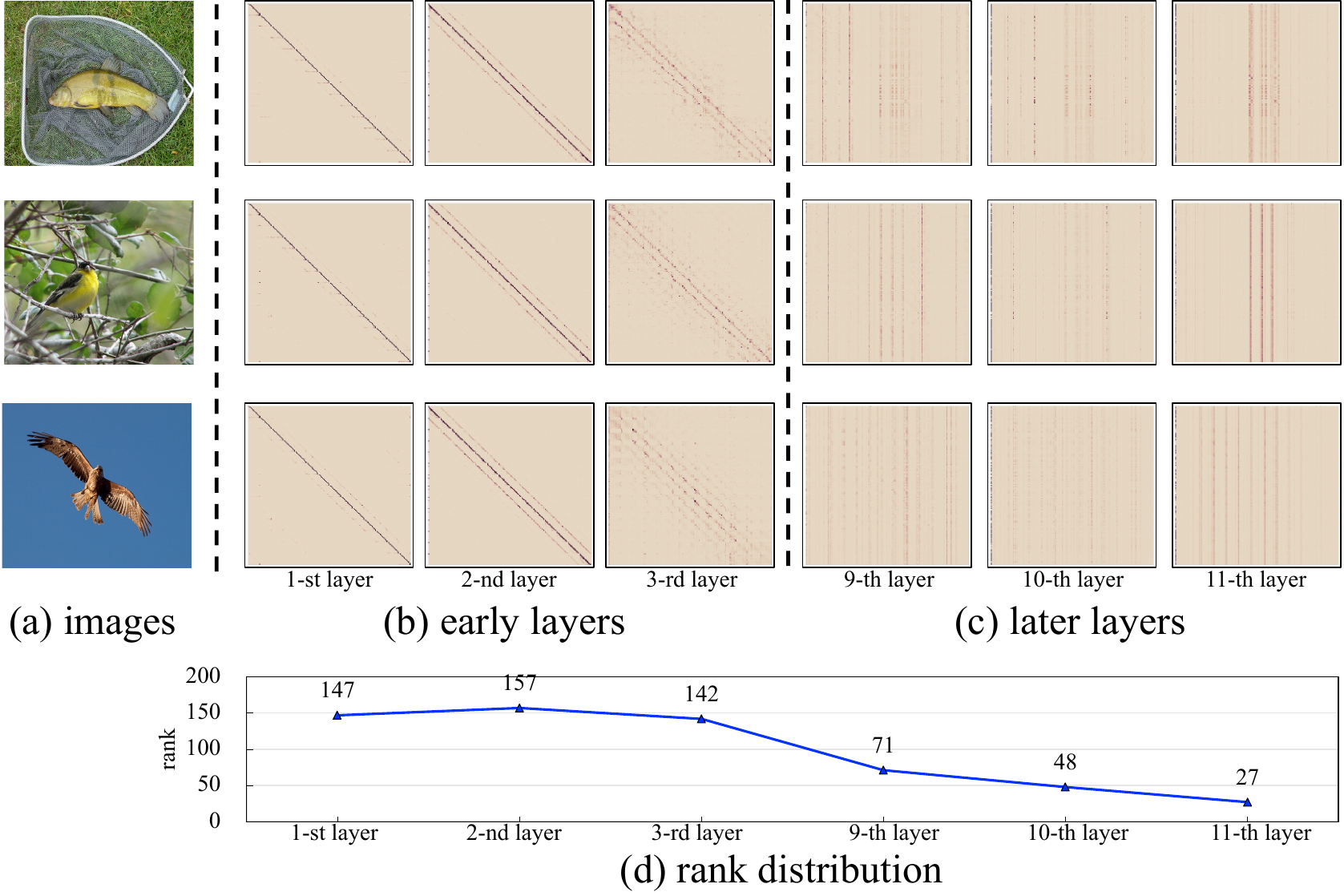} 
   \caption{
   Inhomogeneours attention patterns across layers in ViTs. 
   (a) are three  samples. (b) and (c) are token-by-token attention maps of corresponding heads of (a) in early 1-th, 2-nd and 3-rd layer and later 9-th, 10-th and 11-th layer for Deit-Small, respectively. 
   The row axis represents tokens, and the column axis represents the self-attention similarity coefficients by which the current token attends all tokens in these attention maps. 
   (d) shows the  rank distribution \bp{of the attention maps} across these layers when  eigenvalues are truncated  at 99\% accumulated energy .
%   Early layers has static attention patterns and later layers have dynamic attention patterns.   
%   Sub-figure~(a) shows that three different images have similar attention maps in  the first 1, 2 and 3 layers, while 
%   sub-figure~(b) shows that different tokens of  an image have similar attention coefficients vectors  in the later 9, 10, 11 layers.
% \deli{}
   }
   \vspace{-20pt}
   \label{fig:staic_dynamic}
\end{figure}

\begin{figure*}[htp]
  \centering
%   \fbox{\rule{0pt}{2in} \rule{0.9\linewidth}{0pt}}
  \includegraphics[width=0.9\textwidth]{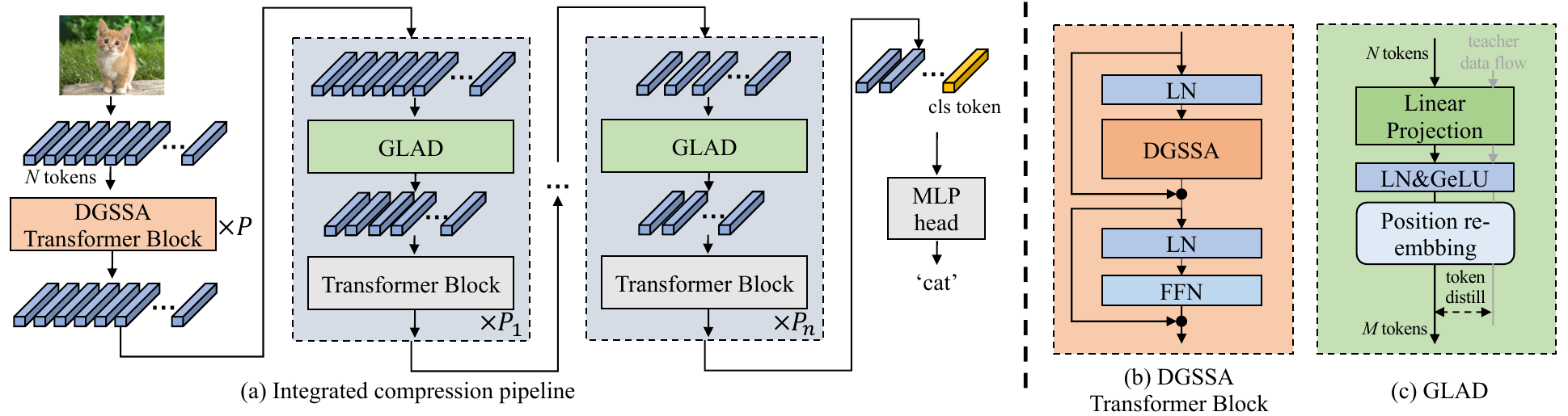}
   \caption{ 
   {The integrated compression pipeline} of dynamic-guided static self-attention~(DGSSA) and global aggregation pyramid~(GLAD). (a) is the integrated pipeline which 
   replaces dynamic self-attention with DGSSA and keeps other modules unchanged in early $P$ layers, and inserts GLAD to reduce the redundant tokens and get a pyramid structure of token numbers in later layers. (b) is the transformer block with DGSSA, and (c) illustrates GLAD. 
   }
   \vspace{-15pt}
   \label{fig:integrated_pipeline}
\end{figure*}

% We observe that
% the attention maps share similar positional patterns among different images in early layers. As shown in Fig~\ref{fig:staic_dynamic}(a), the the first 1, 2 and 3 layers of three different images has similar dynamic attention matrices. 

% The static attention patterns only appears in early layers of ViTs. 

%  \deli{CNNs and Transformers can be thought to be local and global modeling structures, respectively. 
 \deli{ 
%  But dose ViTs have similar attention patterns to Transformers in NLP? } 
Some aforementioned works are motivated by the observed attention patterns of Transformers in NLP, but there are little related explorations for ViTs. 
}
 Therefore, we make observations and find heterogeneous attention patterns in ViTs. 
 As shown in Fig.~\ref{fig:staic_dynamic}, 
 the attention maps \bp{of three different images chosen from ImageNet\cite{deng2009imagenet}} have more similar patterns  in early layers than  later layers, while have more low-rank patterns in later layers than  early layers  \bp{for the well-known ViT model of Deit-Small ~\cite{touvron2021training}}. Obviously, low-rank patterns  indicate the redundancy of tokens.
%  The low-rank patterns reflects token redundancy. 
 
%  it is more suitable to apply static self-attention in early layers than in later layers, as Fig.~\ref{fig:staic_dynamic} shows. 
%  This characteristic of attention pattern can be also \bp{demonstrated} by the findings that  most heads of the early layers pay more  \bp{attention to} position-aware local feature in \cite{cordonnier2019relationship,d2021convit}. 
 
%  \bp{However,} ConViT~\cite{d2021convit} only focus the mixture of local and global attention patterns and misses the mixture of static and dynamic attention.

 However, \deli{MLP-Mixer~\cite{tolstikhin2021mlp} assumes similar patterns in all the layers for their attention computation, which is obviously contradictory to our findings in Fig.~\ref{fig:staic_dynamic}}. 
 %use static attention in all layers, which conficts with the real characteristc of attention patterns of ViTs. Moreover, the  randomly initialized and unguided training of attention matrices may not lead to meaningful patterns. } 
 %\bp{?Even though ?,} their findings can indicate that the early layers have more trend to keep content-free static attention then later layers.
 \bp{In addition, prior} token pyramid methods\cite{tang2021patch,rao2021dynamicvit,xu2021evo,wang2021pyramid,liu2021swin,wu2021cvt,chu2021twins} cannot accelerate the whole ViTs, because
 token redundancy is relatively low and token
 number  is always kept nearly unchanged after compression in early layers, as  Fig.~\ref{fig:staic_dynamic}(d) indicates.

% MLP-Mixer~\cite{tolstikhin2021mlp} rethinks the necessity of dynamic self-attention and replaces it with multilayer perceptrons~(MLPs). Synthesizer~\cite{tay2021synthesizer}  replaces the dot product-based self-attention with random matrices.
% They are  equivalent to using content-free but trainable matrices as importance weights to make global aggregation of features.
% The static attention runs fast and needs no specialized software or hardware framework. However, MLP-mixer and  Synthesizer use static attention in all layers, which conficts with the real characteristc of attention patterns of ViTs. Moreover, the  randomly initialized and unguided training of attention matrices may not lead to meaningful patterns. 

% heretical . 前面是static，；
% 动态性源自视觉语义，；底层特征，高层特征；；；

% Others use static local attention. That is equivalent to a convolutional layer, as  \cite{cordonnier2019relationship} theoretically proves. 
% LeViT~\cite{graham2021levit}  uses pure convolutional layers in the early stages of whole backbone. 
% The local attention pattern introduces hard-coded inductive bias, and thus can benefit data-efficient training, but may limit model capacity in the regime of rich training data. 
% Moreover, they only capture the local-range dependencies, and thus may miss some feature details.  

Inspired by the observed heterogeneous attention patterns, we propose an integrated compression pipeline \bp{that} exploits both  similar patterns across images in early layers and low-rank patterns in later layers \bp{to overcome  the drawbacks of previous works and accelerate a whole  ViT model.} 
As shown in Fig.~\ref{fig:integrated_pipeline}, \bp{this integrated model compression scheme} includes two \bp{key} parts, \bp{ that is,} dynamic-guided static self-attention~(DGSSA) and Global Aggregation Pyramid~(GLAD).

To boost the performance of static self-attention~(SSA), we have SSA inherit some information from the dynamic SA rather than discarding it \bp{away}, like \cite{tolstikhin2021mlp,raganato2020fixed}.
\deli{So  DGSSA is first proposed} \bp{to leverage} the replaced dynamic attention to initialize the trainable attention matrices. 
% During the training phase, we use the dynamic attention to distill the static attention \bp{which can improve the feature representation ability of attention map}. 
Besides, 
motivated by the low-rank patterns,  we propose 
a method of GLAD to reduce token numbers according to rank, which reflects the required minimum number of tokens to represent the attention map. GLAD  can aggregate the full number of tokens into less number of tokens in a \textit{global} scope by a linear projection. \bp{The contributions of this work can be summarized as follows:}

% 1) We \bp{empirically analyzed} the difference in attention patterns across different layers in ViTs, and show that the early layers has static attention patterns, while the later layers has dynamic attention patterns.

1) \bp{A novel} method of DGSSA for early layers \bp{to save computation is proposed}. 
Differently, the static self-attention matrices are initialized with the self-attention information inherited form the dynamic self-attention. 

% The proposed method uses static global attention, which can capture more detailed features than static local attention ~(CNN).  
% Dynamic attention map of pre-trained ViT is used to guide the optimization of the shared static attention map, thus the guided global attention is better than normal global attention~(MLP).

2) We \bp{advance} a \bp{new} method of GLAD to reduce token numbers in later layers.  Different from pruning-based methods and local aggregation methods, GLAD aggregates tokens globally  by a linear projection, which can be more 
compatible to real spatial distribution of token redundancy. 

3) \bp{Extensive} experiments \bp{validate} the effectiveness of DGSSA, GLAD and the integrated pipeline comprising the two methods.  The integrated pipeline can accelerate up to about 121\%  run-time throughput of Deit.

\section{Related Works}
% \subsection{static }
% \subsection{similarity intra attention maps}
% \subsection{}
% Efficient sparse attention methods have been proposed in  NLP tasks~\cite{child2019generating_sparse_transformer,beltagy2020longformer,qiu2019blockwise} under the knowledge of  intrinsic sparsity of self-attention. 

% The mechanism of self-attention makes transformers capture long-range global dependencies, but leads to quadratic computation and memory consumption with respect to the number of tokens. 

%To mitigate the limitation of too much consumed computation and memory of self-attention, its architecture  is modified with a new more efficient form of 
% \textit{dynamic efficient attention}, such as 
%\textit{sparse self-attention} and \textit{low-rank self-attention}, etc, or a form of \textit{static self-attention}, due to characteristic of self-attention. 
%Another line of works keeps the architecture of self-attention unchanged, but prunes some tokens or makes local aggregation of feature to reduce the spatial size of tokens. A structure of  \textit{token pyramid} is obtained in this way. 

% \subsection{Attention Patterns}

\subsection{Efficient self-attention}
To reduce  computation and memory consumed by self-attention, 
its architecture  is modified with a new more efficient form of sparse, low-rank and static versions. 

 \textbf{Sparse self-attention} is adopted by some works to handle long sequence.
 Parmar~\etal~\cite{parmar2018image} restricts the self-attention mechanism to attend  local neighborhoods rather than all positions. Qiu~\etal~\cite{qiu2019blockwise} uses block-wise sparse attention. 
Child~\etal~\cite{child2019generating} uses sparse factorizations of the full attention matrix, and that sparse version makes attending at ﬁxed intervals. Longformer~\cite{beltagy2020longformer} adopts not only that sparse pattern but also a few  additional task-motivated global attention patterns. 
The Reformer~\cite{kitaev2020reformer} uses locality sensing hashing~(LSH) to find and attend the nearest neighbors of
the attention query. 
Routing Transformer~\cite{roy2021efficient} uses online k-means to learn dynamic sparse attention patterns, and thus avoids 
the computation of attending to the unrelated content.
Beyond pure local attention, ETC~\cite{ainslie2020etc} uses a global-local attention mechanism. BigBird~\cite{zaheer2020big} combines global-local attention and random sparse attention, and proves such kind of sparse attention is an universal approximator of sequence functions and Turing
complete. 

\textbf{Low-rank self-attention} is also explored to \bp{yield} efficient self-attention, besides sparsity. 
% the characteristic of attention map is also low-rank, as some works points out. 
Linformer~\cite{wang2020linformer} demonstrates that the self-attention mechanism can be approximated by a low-rank matrix. To encode  fine-grained local
information, Transformer-LS~\cite{zhu2021long} integrated  a novel low-rank projected long-range attention and a local window attention. 

\textbf{{Static self-attention }} can   accelerate 
the self-attention  by  using a static  attention matrix shared by all images 
instead of calculating repeatedly attention matrix for each images. 
%  Different from sparse dynamic attention,  static attention has \bp{faster} run-time speed and needs no specialized software or hardware framework.
MLP-Mixer~\cite{tolstikhin2021mlp} proposes an architecture based exclusively on MLPs, two kinds of which mix the per-location features and spatial information, respectively. The latter mixing MLPs are analogical to trainable static attention matrices. 
% Synthesizer~\cite{tay2021synthesizer} learns synthetic attention weights without token-token interactions, and that attention weights can be random or factorized in NLP tasks. 
Raganato~\etal~\cite{raganato2020fixed} replaces all but one attention head of each encoder layer with static non-trainable attention patterns that are solely based on position in NLP tasks. 
ConViT~\cite{d2021convit} introduces trainable gated combination of standard self-attention and positional self-attention, which is equivalent to a convolutional layer with inductive bias. 
% The trained gate likely chooses positional patterns in early layers, which can support our observation in Fig.~\ref{fig:staic_dynamic} to some degree. 

The proposed method is different from \bp{the above} methods in two aspects. Firstly,  our method only applies static attention for early layers instead of  all layers.
% based on the observed heterogeneous attention patterns of  ViTs in Fig.~\ref{fig:staic_dynamic}.
Secondly, our method initializes the static matrices under the guidance of  dynamic self-attention. 

\subsection{Token Pyramid}
ViTs can be accelerated by pruning or aggregating tokens. Since less tokens remain in the more later layers, there is a structured or unstructured token pyramid. 

\textbf{Pruning-based methods.} Patch Slimming~\cite{tang2021patch} prunes useless patches in a top-down paradigm, where  the effective patches are firstly identified in the last layer and  then used to guide the patch selection  of previous layers. DynamicViT~\cite{rao2021dynamicvit} discards redundant tokens progressively and dynamically based on the importance score, which is provided by lightweight prediction module.  
Evo-ViT~\cite{xu2021evo} updates the selected informative tokens and uninformative tokens with different computation paths, which is called slow-fast token evolution. ATS~\cite{fayyaz2022adaptive} samples informative tokens and discards uninformative tokens from token scoring distribution. TOME~\cite{bolya2022token} prunes tokens by merging two tokens with high similarity after bipartite soft matching. TPS~\cite{wei2023joint} first selects  uninformative tokens, and then squeeze them informative tokens to reduce token numbers. However, 
pruning-based methods will drop spatial information of token sequences, and the lost information is never recovered in later stages. 

% pruning tokens may miss spatial information of token sequence.

\textbf{Aggregation-based methods.} Some works progressively  shrink  the pyramid  of feature maps stage by stage. Then, 
% Token sequence can be reshaped to 2D feature maps. Then,  feature maps are shrunk. 
the shrunk feature maps are flattened into smaller sequence of tokens back.  
Shrinking feature maps can preserve more spatial information than pruning-based methods.
% PVT~\cite{} 
Swin~\cite{liu2021swin} and Twins~\cite{chu2021twins} concatenate 2×2 local neighboring tokens into one and extend feature dim $C$ to  $4C$. 
A following linear layer is applied on the $4C$-dimensional concatenated features. 
CvT~\cite{wu2021cvt} uses down-sampling  with fixed stride to get smaller spatial size of feature maps. 

Unlike these local aggregation-based methods, the proposed GLAD makes  global aggregation of tokens by a linear projection.

\section{Methods}
\label{sec:methods}
% In this section, we propose two methods of dynamic-guided static self-attention~(DGSSA) and global aggregation pyramid~(GLAD)  to compress 
% % the early and later layers of ViTs
%  ViTs 
% % due to the different  characteristic of attention patterns of early and later layers.  
% due to the observed special attention pattern in early layers and the token redundancy in later layers, respectively. 

In this section, we first  review the dynamic self-attention. Then,  
static self-attention  and its  dynamic-guided form are introduced 
to relieve computation of self-attention in early layers.
% and its  dynamic-guided way is formulated.   
Next, 
% considering 
% the intrinsic low-rank characteristic of attention patterns of later layers, 
% the token redundancy, 
% we propose the method of  
% for the sake of both preserving spatial information and,
we introduce GLAD to reduce token numbers for later layers.
Finally, we give an unified framework to integrate both methods in a given ViT.

\subsection{Review of Dynamic Self-Attention}
Self-Attention~(SA) is the key component of Transformers. SA aggregates the sequence features $X\in \mathbb{R}^{N\times d}$ by multiplying its value embeddings $XW_v \in \mathbb{R}^{N\times d}$ with self-attention matrix $A \in \mathbb{R}^{N\times N}$, where $A$ is determined by the the similarity of its query embeddings  $XW_k \in \mathbb{R}^{N\times d}$ and its key embeddings $XW_q \in \mathbb{R}^{N\times d}$ using inner dots. The whole formulation of SA is  as follows:
\begin{equation}
   \left\{
      \begin{array}{c} 
    A = \text{softmax}( \frac{XW_k W_q^TX^T}{\sqrt{d}} ) \in \mathbb{R}^{N \times N} \\
    \text{SA}(X) = A XW_k \in \mathbb{R}^{N \times d},
      \end{array}
   \right .
   \label{eq:sa}
\end{equation}
 where $W_k$, $W_q$, $W_v \in \mathbb{R}^{d \times d}$ are trainable matrices. 
% Features $X$ of each image has its own attention matrix $A$ in general,  
% because its calculation depends on features $X$. 
The process of $\text{SA}$ is conducted like Fig.~\ref{fig:ssa_duibi}(a). It is obvious that the calculation complexity of SA is quadratic to the token numbers $N$, and is unbearable when \bp{facing up} high resolution of images with large $N$. 
% Thus, there are many related research works making efficient SA.

\subsection{Dynamic-Guided Static Self-Attention}
In this \bp{part}, to relieve the computation of SA, we  introduce the static self-attention~(SSA)  applied in the early layers of ViTs. Then, we discuss the drawbacks of two existing forms of SSA. Finally, a new type of SSA is proposed. 
% Finally, we propose a new method to get optimal static attention matrices.

\subsubsection{Static Self-Attention}
We observe that the different images share similar self-attention matrix $A$ in early layers of ViTs shown in Fig.\ref{fig:staic_dynamic}~(b), despite the dynamic manner \bp{in nature}. 
% In general, early 
% layers have \bp{?content-free attention?} patterns. Thus, 
Assuming there is static self-attention matrix $\hat{A}$, which is an optimal estimation of $A$. 
We use it to replace $A$ in Eq.\ref{eq:sa}, and obtain the static self-attention~(SSA), as follows:
\begin{equation}
    \text{SSA}(X) = \hat{A} XW_k \in \mathbb{R}^{N \times d}.
    \label{eq:ssa}
\end{equation}
In this way, the calculation  \bp{may be more} efficient due to getting \bp{rid} of the quadratic complexity of attention matrix $A$. 
The characteristic of attention patterns makes 
this simplification plausible, and thus the application of efficient SSA \bp{may} not cause much accuracy drop in early layers. 

\subsubsection{Dynamic-guided type}
\begin{figure}[htp]
  \centering
%   \fbox{\rule{0pt}{2in} \rule{0.9\linewidth}{0pt}}
  \includegraphics[width=0.8\linewidth]{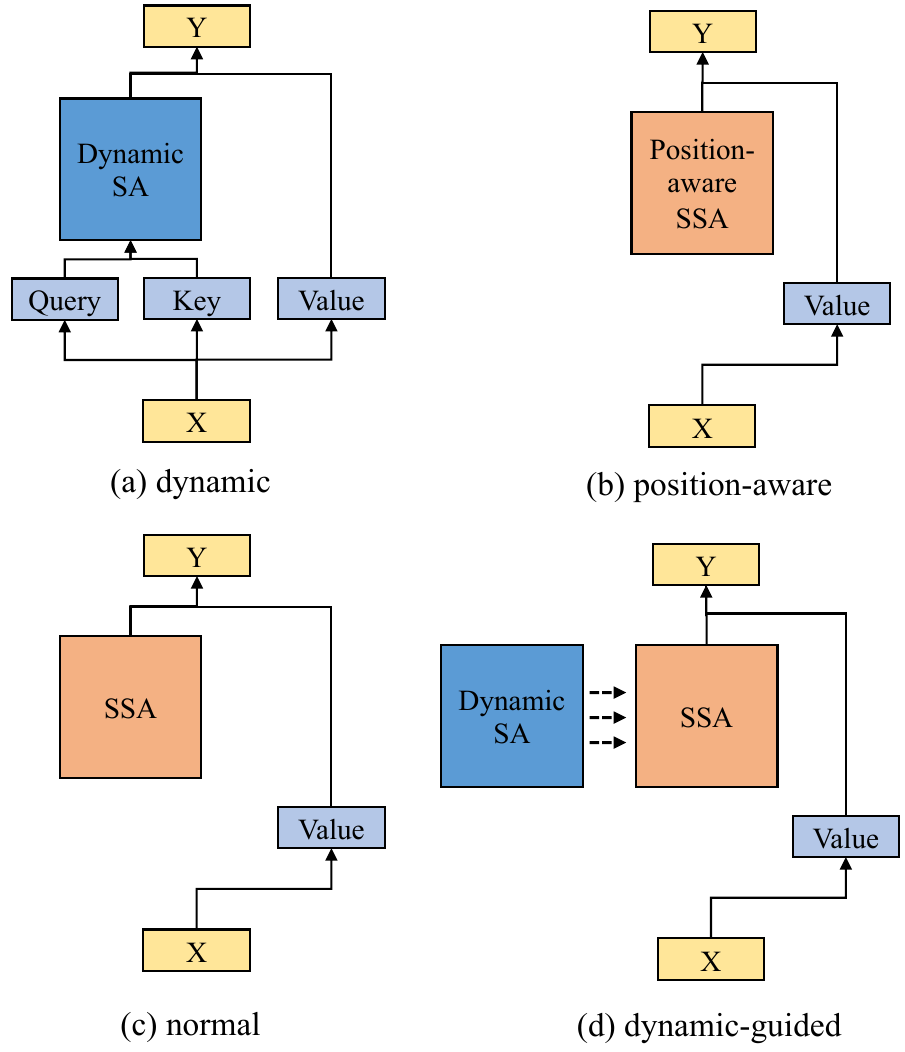}
   \caption{ 
   \deli{Dynamic self-attention~(SA) and three types of static self-attention~(SSA). (a) denotes dynamic SA, 
(b) denotes position-aware SSA used by ~\cite{cordonnier2019relationship,d2021convit},
  (c) denotes normal SSA and  
    (d) denotes the proposed DGSSA method, which inherits information from the dynamic SA.}
   }
   \vspace{-10pt}
   \label{fig:ssa_duibi}
\end{figure}

% \begin{figure}[htp]
%   \centering
% %   \fbox{\rule{0pt}{2in} \rule{0.9\linewidth}{0pt}}
%   \includegraphics[width=0.8\linewidth]{figs/cvpr-img-ssa.pdf} \hspace{20pt}
%   \caption{ 
%   The illustration of dynamic-guided training of static self-attention.  The dash line denotes the data flow which only occurs in training phase. 
%   }
%   \label{fig:ssa}
% \end{figure}
We review the two existing types of position-aware  and  normal SSA, as illustrated in Fig.~\ref{fig:ssa_duibi}(b) and (c), respectively. Then, we introduce the proposed dynamic-guided SSA.

% In this part, we first analyse the drawbacks of the two types and then introduce the proposed dynamic-guided SSA~(DGSSA).

\textbf{Position-aware SSA. } 
The position-aware attention pattern comes to one's mind at first glance at  Fig.~\ref{fig:staic_dynamic}(a).
Based on this idea, position-aware SSA~ \cite{cordonnier2019relationship,d2021convit} makes aggregation of value embeddings by a diagonal matrix $\hat{A}$, as shown in Fig.~\ref{fig:ssa_duibi}~(b). 
 Actually, this type of static self-attention  
is equivalent  to a convolutional layer \cite{cordonnier2019relationship}. 
However,  it is not reasonable to assume self-attention has completely local position-aware patterns, because 
 some global details will be ignored.
We find that this type is  inferior to the normal \bp{one} and the proposed dynamic-guided  in comparison experiments, \bp{which will be shown} in Tab.~\ref{tab:comparison_attention_1} and \ref{tab:comparison_attention_2}. 

\textbf{Normal SSA}. 
A full static attention matrix $\hat{A}$ is used by normal SSA to make global aggregation of value embeddings, as shown in Fig.~\ref{fig:ssa_duibi}(c), thus the normal type can capture more details than the position-aware type. That explains why ``normal'' outperforms ``local'' in most settings in Tab.~\ref{tab:comparison_attention_1} and \ref{tab:comparison_attention_2}. 
% MLP-Mixer~\cite{tolstikhin2021mlp} also uses a MLP layer to replace self-attention layer in Transformers. 
However, normal SSA still has inferior performance to dynamic self-attention,  to which we attribute that the learned attention matrix has less useful patterns due to meaningless initialization.
% We visualize the patterns learned by a randomly initialized MLP layer, which is denoted as ``normal'',  and by a MLP layer initialized by our proposed method , which is denoted as ``dynamic-guided'', as shown in Fig.~\ref{fig:learned_attention}. 

\textbf{Dynamic-guided SSA.}
Considering the drawbacks of two types of static self-attention, we propose a method of dynamic-guided static self-attention. 
Following normal SSA, we also use a full matrix $\hat{A}$. Instead of training $\hat{A}$ from scratch, we use 
 dynamic attention matrix $\hat{A}$ to initialize it. 
  As shown in Fig.~\ref{fig:ssa_duibi}(d),  the dynamic-guided SSA can inherit self-attention information from the replaced dynamic SA rather than discarding it, like normal SSA. 
 The performance comparison of the two types validates our claim, as shown in Tab.~\ref{tab:comparison_attention_1} and \ref{tab:comparison_attention_2}.

 In our opinion, the aggregated value embeddings by the static self-attention should mimic these obtained by dynamic self-attention, and thus the optimization problem is formulated to determine the optimal $\hat{A}$, as follows: 
\begin{equation}
    \min_{\hat{A}} \mathbb{E}_{X}[||(\hat{A}-A)XW_k||^2], 
    \label{eq:ssa_loss}
\end{equation}
where $||\cdot||$ is \bp{the $L_{2}$}  norm. We derive its closed-formed solution as 
\begin{equation}
    \hat{A} = \mathbb{E}_{X}[AXW_kW_k^TX^T](\mathbb{E}_{X}[XW_kW_k^TX^T])^{-1}.
    \label{eq:closed_form_solution}
\end{equation}
Then, the optimal estimation $\hat{A}$ can serve as meaningful initialization  for the following training of the compressed ViT.
% In some special settings, static self-attention of $\hat{A}$ has negligible accuracy drop compared with dynamic self-attention. Though,  the following fine-tuning is also recommended in most settings.  

% Then estimated $\hat{A}$ serves as the initialization for the fine-tuning. 

\subsection{Global Aggregation Pyramid }

\begin{figure}[tp]
  \centering
%   \fbox{\rule{0pt}{2in} \rule{0.9\linewidth}{0pt}}
  \includegraphics[width=\linewidth]{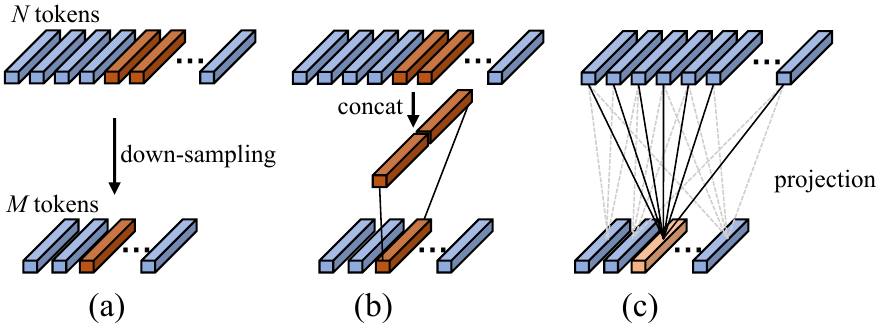}
   \caption{ 
   Different token aggregation methods of reducing $N$ tokens to $M$ tokens. (a) and (b) denote local down-sampling and concatenation method, respectively, while (c) denotes the proposed GLAD method, which makes global aggregation to reduce token numbers. 
   }
   \label{fig:GLAD_duibi} 
   \vspace{-15pt}
\end{figure}

% \begin{figure}[htp]
%   \centering
% %   \fbox{\rule{0pt}{2in} \rule{0.9\linewidth}{0pt}}
%   \includegraphics[width=0.5\linewidth]{figs/cvpr-img-tpyd.pdf}
%   \caption{ 
%   The illustration of token pyramid method.  The black line denotes the data flow of compress ViTs, while the grey line denotes the data flow of teacher ViTs. 
%   }
%   \label{fig:ssa}
% \end{figure}
To exploit the low-rank attention patterns in later layers, 
% Tokens have similar attention vectors in later layers, 
as shown in Fig.~\ref{fig:staic_dynamic}(b),
% , because many tokens attend the same semantic region. Thus, there are quite much \bp{redundancy among} tokens in later layers. 
% These redundant tokens can be removed by token pruning methods, which may miss some spatial information. Unlike token pruning methods, aggregation-based methods, such as   down-sampling with stride and concatenation in Fig~\ref{fig:GLAD_duibi}(a) and (b) respectively, aggregates locally some adjacent tokens into one in a fixed-size window and can also decrease the number of tokens. However, this fixed-size window may not be suitable for the real distribution of feature information. 
 we propose a method of global aggregation pyramid ~(GLAD)  \bp{to reduce the {token} redundancy } 
 by aggregating the full number of tokens into less number.
The existing local aggregation methods make down-sampling and concatenation in a fixed  scope to reduce token numbers $N$ to $M$, as shown in Fig.~\ref{fig:GLAD_duibi}(a) and (b), respectively. The proposed GLAD can  aggregate tokens in a global scope, as shown in Fig.~\ref{fig:GLAD_duibi}(c). 

% \deli{ Fig.~\ref{fig:GLAD_duibi} illustrates the difference between the proposed linear projection method and  other two aggregation-based methods. }
 
 The architecture of GLAD is shown in Fig.~\ref{fig:integrated_pipeline}(c). Its key component consists \bp{of} a linear projection $Y=WX$,  where $W \in \mathbb{R} ^{M \times N}$ is a trainable weight and $M < N$. The projection  globally aggregates a token sequence $X \in \mathbb{R} ^{N \times d}$, whose length is $N$, into  a new sequence $Y\in \mathbb{R} ^{M \times d}$, whose length is $M$.
 
\textbf{LayerNorm and GeLU.} 
 Since LayerNorm and GeLU can stabilize the training and introduce additional non-linearity, respectively, 
% Since the pure linear projection without BatchNorm and GeLU makes the training diverge, \deli{as we find in Tab.~\ref{tab:comparison_GLAD}},
we add them after the linear projection. 
The total formulation of GLAD is shown below: 
 \begin{equation}
%  \text{GLAD}(X, M) = \text{GeLU}( \text{BatchNorm}(WX )). 
 Y = \text{GeLU}( \text{BatchNorm}(WX )). 
 \label{eq:convbngelu_forseqlen}
 \end{equation}
 
    %  Position re-embedding makes the spatial information of transformed features distinguished. Token distill loss helps the training of the compressed ViT. 

\textbf{Position re-embedding.} 
% There are one issue on the transformed feature $Y$.  The spatial information of $Y$ is  missing.  
The original feature sequence $X$ is position-aware due to the added position embedding  before the first layer~, as \cite{touvron2021training} does.  However, the linear projection of $X$ is spatially position-free. Thus, the transformed feature $Y$ will lose spatial information. To mitigate this problem, we add a new position embedding matrix $E\in \mathbb{R}^{M\times d}$ on $Y$ to  re-encode  the position information. Thus,  the Eq.\ref{eq:convbngelu_forseqlen} can be re-written as:
 \begin{equation}
%  \text{GLAD}(X, M) = \text{GeLU}( \text{BatchNorm}(WX ))+ E.   
 Y = \text{GeLU}( \text{BatchNorm}(WX ))+ E.   
 \label{eq:convpositionbngelu_forseqlen}
 \end{equation}

\textbf{Token distillation.} There is always a pre-trained model, denoted as $\text{vit}_{\text{pre}}$, in the scenario of model compression. 
 \bp{We intend to use $\text{vit}_{\text{pre}}$ to help the training of compressed ViTs by token distillation on the training dataset $\mathcal{D}$.} Specifically, 
we use the token features $X_t$ of $\text{vit}_{\text{pre}}$ to distill the features $Y$ obtained by GLAD. However, $X_t \in \mathbb{R}^{N \times d}$ and $Y \in \mathbb{R}^{M \times d}$ have different sequence length.  To solve this problem, we use another instance of GLAD to transform  $X_t$ to obtain $Y_t$, where $Y_t \in \mathbb{R}^{M \times d}$. 
The grey line denotes the data flow of $\text{vit}_{\text{pre}}$ in Fig.~\ref{fig:integrated_pipeline}(c). 
The token distillation loss is defined as the MSE between features $Y$  and $Y_t$ 
% \deli{before the position re-embedding module}, 
, as shown below: 

\begin{equation}
      \begin{array}{c} 
    f(M;\text{vit}_{\text{pre}},\mathcal{D} ) = ||Y-Y_t||^2. 
      \end{array}
   \label{eq:glad_loss}
\end{equation}
 
\subsection{Integrated Compression Pipeline}
\label{sec:integrated_compression_pipeline}

% SSA-based compression is limited to the early layers of ViTs, while GLAD-based compression always reduces token numbers in later layers and leave early layers nearly unchanged. Thus, it is intuitive to integrate both to produce a better compressed ViT. 
\bp{Based on the above two modules, we integrate  DGSSA and GLAD to effectively} compress a whole ViT. 
We insert them into a $L$-layer ViT which has $n_l$ tokens in the $l$-th layer and has embedding dim of $d$, and the consecutive blocks of the compressed ViT are computed as:
\begin{equation}
      \begin{array}{l} 
    \textbf{x}^{l} = \gamma_l\text{GLAD}(\textbf{x}^{l}, m_l) + (1-\gamma_l) \textbf{x}^{l}\\ 
    \hat{\textbf{x}}^{l} = (1-\phi_l)\text{SA}(\text{LN}(\textbf{x}^{l}))  + \phi_l\text{DGSSA}(\text{LN}(\textbf{x}^{l})) + \textbf{x}^{l}\\
    \textbf{x}^{l+1} = \text{MLP}(\text{LN}(\hat{\textbf{x}}^{l})) + \hat{\textbf{x}}^{l}  \\
      \end{array}
\end{equation}
where binary variables $\gamma_l$ and $\phi_l$ 
% are , and $\phi_l=1$   denotes that 
% GLAD decreases token numbers to $m_l$ in the $l$-th layer,
denotes whether 
DGSSA and GLAD is applied in the $l$-th layer, respectively.
% otherwise GLAD is not applied.  
% $\gamma_l=1$ denotes that SSA is conducted in the $l$-th layer, otherwise standard SA is conducted. 
$m_l \in [1, n_l]$  is  integer variable  and  denotes the  number of left tokens in $l$-th layer.  
Variables $\bm{\gamma}$, $\bm{\phi}$, $\bm{m} \in \mathbb{R}^{L}$ can determine the architecture of the compressed ViT.
 Since \bp{their total combination possibility} is too huge 
 and there may be some unreasonable architectures, we make three constraints on them.  

\textbf{Constraint of $\bm{\gamma}$.} DGSSA is consecutively applied  in the \bp{early} layers~(except the 0-th layer), and GLAD is excluded. For easy illustration, we introduce another scalar variable $P \in [1,L-1]$, which denotes that the first $P$ layers use the SSA. Thus, there are two  constraints of $\sum_{l=1}^{P}\gamma_l = P$ and $\sum_{l=1}^{P}\gamma_l*\phi_l=0$. 
% $\bm{m}$ and $\bm{\phi}$ 

\textbf{Constraint of $\bm{m}$ and $\bm{\phi}$ .} We also introduce a hyper-parameter $Q$ to control the number of layers using GLAD. The constraint is formulated as 
$\sum_{l=P+1}^{L}\phi_{l}=Q$. 
% If $\phi_{l_1}=1$, the token numbers will be changed as $m_{l_1}$ in all layers behind ${l_1}$. Furthermore, if  $\phi_{l_2}=1$, where ${l_1}<{l_2}$, the token numbers will be modified as  $m_{l_2}$ in all layers behind ${l_2}$. 
And there is an implicit constraint of  $m_{l_1} \geq m_{l_2}$, where  ${l_1}<{l_2}$, if $\phi_{l_1}=1$ and $\phi_{l_2}=1$. 
\bp{Thus, later layers have less left token numbers than early layers, and $\bm{m}$ has a shrunk pyramid distribution across layers}. That distribution is similar with pyramid structure of main-stream CNNs.

\textbf{FLOPs Constraint.} Another constraint is related to the compression ratio of  \deli{FLOPs}.  The \bp{FLOPs} of dynamic SA is $2n_l^2d + 4n_ld^2$, and the FLOPs of FFN is $8n_ld^2$ under the mlp ratio of 4. Thus, the total FLOPs of a ViT is $\sum_{l=0}^{L-1}{12n_ld^2+4n_ld^2}$. The FLOPs of DGSSA is $m_l^2d + 2m_ld^2$.
% , and the FLOPs of MLP is $8m_ld^2$. 
Thus, the total FLOPs of a compressed ViT is $\sum_{l=0}^{L-1}{m_l^2d + 10m_ld^2}$. 
% It is notable that although the complexity of SSA is also quadratic to  $m_l$, the run-time speed of SSA is much faster than SA because the calculation of  dynamic attention matrix is eliminated \bp{in our proposed hybrid compression scheme}. 
We compress the original ViT under of FLOPs constraint of  $\frac{\sum_{l=0}^{L-1}{ 10m_ld^2 + m_l^2d }}{\sum_{l=0}^{L-1}{12n_ld^2+4n_ld^2}} <= \eta$, where $\eta$ is compression ratio.

% dd

% dd

% dd

\textbf{Loss formulation.} The \bp{total} loss $\mathcal{L}$  comprising 
the task loss $\mathcal{L}_{\text{task}}$ and the GLAD loss $\mathcal{L}_{\text{glad}}$ is 
optimized with respect to the model weights of compressed ViT and its architecture variables under these constraints.
\bp{The} task loss, such as cross-entropy loss of classification tasks, is denoted as $\mathcal{L}_{\text{task}}(\mathcal{W}, \hat{A}; \bm{\gamma}, \bm{\phi}, \bm{m},  \mathcal{D})$, where  $\mathcal{W}$ and $\hat{A}$ denotes the model weights and static self-attention matrices, respectively.   
% Besides the task loss, we also optimize SSA loss under which dynamic attention guides the training of static attention $A_l$, as shown in Eq.~\ref{eq:ssa_loss}. The calculation of SSA loss involves a non-grad original ViT model,  which provides the dynamic attention and is denoted by $\text{vit}_{\text{pre}}$. 
% DGSSA loss is taken from Eq.~\ref{?} and can be denoted as $loss_{dgssa}(A_l, \gamma_l; \text{vit}_{\text{pre}}, \text{X}, \text{Y})$.
The GLAD loss formulated as $\mathcal{L}_{\text{glad}} =  \sum_{l=P+1}^{L-1} \phi_l \cdot f(m_l; \text{vit}_{\text{pre}}, \mathcal{D})$ based on token distillation loss taken from Eq.~\ref{eq:glad_loss}.
% The total loss function comprising task loss and SSA loss can be defined as \begin{equation}
% \begin{aligned}
%       &\text{Loss}(W, \gamma_l, \phi_l, m_l, A_l;  \text{M}_{\text{vit}}, {\text{X},\text{Y}}) \\
%       =&   loss_{task}(W, \gamma_l, \phi_l, m_l;  {\text{X},\text{Y}})  \\
%       +&loss_{ssa}(A_l; \text{M}_{\text{vit}}, \text{X}, \text{Y}) \\
%       +&loss_{glad}(m_l, \phi_l; \text{M}_{\text{vit}}, \text{X}, \text{Y}) \\
% \end{aligned}
% \end{equation}
% \bp{As such, the joint loss can be }  defined as $loss=loss_{task} + loss_{dgssa} + loss_{glad}$. 
% The total loss $\mathcal{L}$ can be defined as $\mathcal{L}_{\text{task}} + \mathcal{L}_{\text{glad}}$.
In addition, we formulate a constrained-optimization problem to determine the optimal $\bm{\gamma}$, $\bm{\phi}$ and $\bm{m}$ under hyper-parameters of  $P$, $Q$ and $\eta$, as follows:
\begin{equation}
\begin{aligned}
    \min_{\mathcal{W}, \hat{A}, \bm{\gamma}, \bm{\phi}, \bm{m} }& {\mathcal{L}}(\mathcal{W}, \hat{A}, \bm{\gamma}, \bm{\phi}, \bm{m};  \text{vit}_{\text{pre}},\mathcal{D}) \\
    % (W, \gamma_l, \phi_l, m_l;  {\text{X},\text{Y}}) \\
    \text{s.t.}& \sum_{l=1}^{P}\gamma_l = P, \sum_{l=1}^{P}\gamma_l*\phi_l=0,  \gamma_0 = 0\\
    & \sum_{l=P+1}^{L-1}\phi_l=Q , m_{l_1} \leq m_{l_2}, \text{where} \; \; {l_1}<{l_2} \\ 
    & \frac{\sum_{l=0}^{L-1}{10m_ld^2 + m_l^2d }}{\sum_{l=0}^{L-1}{12n_ld^2+4n_ld^2}} <= \eta\\
\end{aligned}
\label{eq:unified_loss}
\end{equation}

\textbf{Decoupling optimization.} Considering the joint optimization of $\mathcal{L}$ with respect to weights and architecture variables is \bp{very hard}, we decouple them. Firstly,  $\bm{\gamma}$ can be determined under the constraint of  $\bm{\gamma}$ under hyper-parameter $P$.
Then, we design an Accuracy Metric~(AM) to determine the optimal   $\bm{m}$ and $\bm{\phi}$ by the constraints of $\bm{m}$, $\bm{\phi}$ and FLOPs under hyper-parameters $Q$ and $\eta$, as follows:
\begin{equation} \label{eq:am}
\begin{split} 
    % \adddotsbeforeeqnnum
    \text{AM}(\bm{m}) = \prod_l^L \frac{\sum_i^{m_l}\sigma_l,i}{\sum_i^{n_l}\sigma_l,i}, 
\end{split}    
\end{equation}
where $\sigma_{l,i}$ denotes the $i$-th expectation of singular value of self-attention matrix to the training samples at $l$-th layer. 
The optimal $\bm{m}$ and $\bm{\phi}$  are obtained by  maximizing AM. We also give a dynamic programming solution in supplementary file for saving pages. 
% when meeting  $\gamma_l$, $\phi_l$ and $m_l$, we leave it as future works and \bp{only} use heuristic rules to determine the value of $\gamma_l$, $\phi_l$ and $m_l$ in Sec.~\ref{sec:Expe}. 
Finally, we optimize the $\mathcal{L}$ with respect to weights $\mathcal{W}$ and $\hat{A}$ by classical gradient descending method after determining the architecture variables.

\section{Experiments}
\label{sec:Expe}
% We conduct experiments 
\subsection{Implementation Details}

\textbf{Datasets.} We conduct compression experiments on ILSVRC-2012~\cite{deng2009imagenet}, which has 1.28M colored training images and 50K colored validation images from 1K classes. 
The top-1 accuracy on a single crop is reported.

\textbf{Training settings. }
The training settings mainly follow Deit~\cite{touvron2021training} and Swin~\cite{liu2021swin}. 
We use an AdamW~\cite{kingma2014adam} optimizer with a learning rate initialized as 0.001 and decayed by $1e^{-5}$ with the cosine strategy. The weight decay of 0.05 is adopted. 
The training lasts for 300 epochs with a linear warm-up  used in the first 5 epochs, and the learning rate of warm-up is $1e^{-6}$.
We also use the same data augmentation and regularization settings as Deit. All models are trained  on  NVIDIA V100 8 GPUs with  a batch size of 1024 for DGSSA-based compression and a batch size of  \bp{ 4096} for the GLAD-based  and integrated pipeline compression. 
Following \cite{rao2021dynamicvit}, we adopt self-supervised distill training. Specifically, we optimize cross-entropy loss to minimize the difference between predictions of a original pre-trained ViT and the compressed one. 

% The features are considered to be the most discriminative before classifiers. Thus, we use the features $o_t \in \mathbb{R}^{d} $ of the pre-trained ViT $\text{vit}_{\text{pre}}$ to distill the features $o \in \mathbb{R}^{d} $ of the compressed ViT. 

\textbf{{Compression configurations.}}
% We apply DGSSA in the first $P$ layers while apply GLAD in the later remaining layers $Q$ times. 
 1) Configurations of Deit. To achieve around 100\% increased throughput, compression ratio $\eta$ of Deit-Small an Deit-Base is set to be around 0.5. The $\eta$ of Deit-Tiny is 0.54, since Deit-Tiny is more compact.
 The hyper-parameter $P$ is  2 based on the empirical experiment results in  Tab.~\ref{tab:comparison_attention_1}.
 Too large $Q$ will introduce much additional overhead computation of GLAD, while too small $Q$ will lead to bad compression results. Thus,
 we empirically set  hyper-parameter $Q$ as 3. 
 ~~2) Configurations of Swin. The compression ratio $\eta$ is set to be 0.1, 
 since Swin already has been quite efficient. 
 The hyper-parameter $P$ is set to be 6, 12 and 12 in the compression configurations of Swin-Tiny, Swin-Small and Swin-Base, respectively. 
Above all, 
$\gamma_l$ is determined under  $P$, while $\phi_l$ and $m_l$ can be determined by maximizing AM in Eq.\ref{eq:am} under $Q$ and $\eta$.  All configurations can be found in Tab.~\ref{tab:configurations}.

% \deli{The DGSSA and GLAD can be used separately, which are called SSA-based compression and GLAD-based compression, respectively. The training settings of single usage are consistent with the above.}
% For the first $P$ layers, 
% we first get the estimated static attention matrix by Eq.~\ref{eq:closed_form_solution}, and then use it as the initialization of $A_l$. Next, we train the compressed ViT from its original pre-trained models under the unified framework of the loss defined in Eq.\ref{eq:unified_loss}.

Taking Our-Deit-S in Tab.~\ref{tab:configurations} as an example to interpret the configuration. DGSSA is applied at the 1-th and 2-nd layer. 
% GLAD is applied 3 times in the later remaining layers. To be specific, 
The token numbers 197 is reduced to 127, 77 and 35 when GLAD is applied at 3-th, 5-th and 7-th layer.
% We give an example to illustrate how to interpret  $\gamma_l$, $\phi_l$ and $m_l$. 
% If we have $P=3$, then $\gamma_1$, $\gamma_2$ and  $\gamma_3$ are 1. 
% If we have $q=3$ and $\phi_3=1, m_3=70$, $\phi_5=1, m_5=50$, $\phi_7=1, m_7=42$. We can infer the complete $m$ as [197, 197, 197, 70, 70, 50, 50, 42, 42, 42, 42, 42]. 

% , except for the number of training epoch. The training of SSA needs 100 epochs, while the training of GLAD still needs 300 epochs.

\subsection{Results on ILSVRC-2012}
\begin{table*}[h]
  \begin{center}
      \caption{Comparison of compressed ViTs with previous methods on ILSVRC-2012. Following \cite{touvron2021training,liu2021swin}, we measure throughput based on the public code  on a V100 GPU.
      $^*$ means that 428.8 is our measured speed while 436.9 is the value provide by \cite{liu2021swin}. 
      }
    %   \vspace{-12pt}
      \label{tab:benchmark}
      \footnotesize
  \begin{tabular}{|c|l|c|c|c|c|c|}
  \hline
  \multirow{2}{*}{Model} & \multirow{2}{*}{Method} &\multirow{2}{*}{Param} & \multirow{2}{*}{FLOPs}  & \multicolumn{2}{c|}{Throughput}  & \multirow{2}{*}{Top-1 Acc. (\%)}  \\
\cline{5-6}
  &&&& (img/s)& $\uparrow$(\%) & \\
  \hline
  % Su and Lu &-&&&& 
  \multirow{13}{*}{\makecell[c]{ Deit}}
  &Deit-Tiny\cite{touvron2021training}~(baseline)&  5.7M &1.3G&2536&-&72.2\\
  &S$^2$ViTE~\cite{chen2021chasing} & 4.2M & 1.0G & -& 11.8 & 70.1 \\ 
  &Evo-Deit-T\cite{xu2021evo} &-  &- & 4027& 58.8 &72.0 \\
  &PS-Deit-T\cite{tang2021patch}  &-  &0.7G & - & 68.1 &72.0 \\
    % &DynamicViT-Deit-T\cite{rao2021dynamicvit} & - & - & 3890 & 53.4 & 71.2 \\
%   &GLAD-Deit-T &5.8M  &0.7G & 4670& 84.1  &-  \\
%   &Integrated-Deit-T(ours) &5.7M  &0.6G & 4767& \textbf{88.0}  &71.8  \\
  &TPS\cite{wei2023joint}& 5.9M & 0.8G & - & - & 72.9 \\
  &Integrated-Deit-T(ours) &5.7M  &0.7G & 4410& \textbf{73.9}  &72.1  \\
  \cline{2-7}
  &Deit-Small\cite{touvron2021training}~(baseline)& 22.1M &4.6G & 940&- &79.9 \\
  &S$^2$ViTE~\cite{chen2021chasing} & 14.6M & 3.1G & -& 29.3 & 79.2 \\ 
  &IA-RED$^2$~\cite{pan2021ia} & - & - & -& 46.2 & 79.1 \\ 
    &Evo-Deit-S\cite{xu2021evo} &-  &- & 1510& 60.6 &79.4 \\
  &PS-Deit-S\cite{tang2021patch}  &-  &2.6G & -& 64.5 &79.4 \\
  &DynamicViT-Deit-S\cite{rao2021dynamicvit} & - & 4.0G & 1525 & 62.2 & 79.8 \\
  &ATS\cite{fayyaz2022adaptive} & 22.1M & 2.9G & - & - & 79.7 \\
  &TPS\cite{wei2023joint}& 22.1M & 3.0G  & - & - & 80.1 \\
  &TOME\cite{bolya2022token}& - & 2.7G & 1550 & 64.9 & 79.4 \\
  &GLAD-Deit-S(ours) &22.2M  &2.3G & 1934& 106.0 &79.9 \\
  &Integrated-Deit-S(ours) &21.6M  & \textbf{2.1G}& \textbf{2080}& \textbf{121.1} 
  &79.9 \\
\cline{2-7}
  &Deit-Base\cite{touvron2021training}(baseline)& 86.6M &17.5G & 292&- &81.8 \\
%   &S$^2$ViTE & 56.8M & 11.7G & -& 32.8 & 82.2 \\ 
  &IA-RED$^2$~\cite{pan2021ia} & - & - & -& 37.5 & 80.9 \\ 
    &Evo-Deit-B\cite{xu2021evo} &-  &- & 462& 54.5 &81.3 \\
  &PS-Deit-B\cite{tang2021patch}  &-  &9.8G & - & 67.8 &81.5 \\
    % &DynamicViT-Deit-B\cite{rao2021dynamicvit} & - & - & 454 & 51.8 & 80.8 \\
%   &GLAD-Deit-B &86.8M  &9.1G & 585& \textbf{100.3}  &-  \\
  &Integrated-Deit-B(ours) &84.5M &8.6G & \textbf{622}& \textbf{113.0}  &\textbf{81.6}  \\
  \hline
  \multirow{6}{*}{\makecell[c]{ Swin}}
  &Swin-T\cite{liu2021swin}(baseline) &29.0M &4.5G& 755.2&- &81.3 \\
  &DGSSA-Swin-T(ours) &28.5M  & 4.1G & 823 & 9.0 & 81.3 \\ 
 \cline{2-7}
  &Swin-S\cite{liu2021swin}(baseline) &49.6M &8.7G& 428.8(436.9)$^*$&- &83.0 \\
  &DGSSA-Swin-S(ours) &48.7M  &8.0G & 474 & 10.5 &\deli{83.0} \\
  \cline{2-7}
  &Swin-B\cite{liu2021swin}(baseline) &87.8M &15.4G& 278.1&- &83.5 \\
  &DGSSA-Swin-B(ours) &85.4M  &14.1G & 306 & 10.0 & 83.4 \\
  \hline
  \end{tabular}
%   \footnotetext{428.8 is our measured speed while 436.9 is the value provide by \cite{liu2021swin}}
  \end{center}
%   \vspace{-8mm}
\vspace{-20pt}
\end{table*}

\begin{table}[htp] 
  \begin{center}
  \scriptsize
  \caption{
  Configurations of our compressed ViTs in Tab.~\ref{tab:benchmark}}
  \label{tab:configurations}
  \begin{tabular}{|c|c|c|c|c|}
   \hline
\multirow{2}{*}{}& {SSA} & \multicolumn{3}{c|}{GLAD} \\  
\cline{2-5}
       &$P$ & $Q$ & $\bm{\phi}$ & $\bm{m}$ \\
   \hline
Integrated-Deit-T & 2 & 3 & $\phi_3$, $\phi_5$, $\phi_7$ & 197 $\rightarrow$ 152 $\rightarrow$ 107 $\rightarrow$ 62 \\
Integrated-Deit-S & 2 &3& $\phi_3$, $\phi_5$, $\phi_7$ & 197 $\rightarrow$ 127 $\rightarrow$ 77 $\rightarrow$ 35 \\
Integrated-Deit-B & 2 &3& $\phi_3$, $\phi_5$, $\phi_7$ & 197 $\rightarrow$ 143 $\rightarrow$ 89 $\rightarrow$ 35 \\
DGSSA-Swin-T & 6 & - & - & - \\
DGSSA-Swin-S & 12 & - & - & - \\
DGSSA-Swin-B & 12  & - & - & - \\
\hline
  \end{tabular}
  \end{center}
  \vspace{-12pt}
\end{table}

In this section, we compress two typical ViTs of Deit~\cite{touvron2021training} and Swin~\cite{liu2021swin} on ILSVRC-2012. 
% Deit uses global attention and has equal token numbers in each layer, while Swin uses local window attention and has a hierarchical distribution of token numbers across layer. 
We use the integrated pipeline of DGSSA and GLAD to compress Deit-Tiny, Small and Base, and the results are denoted as Integrated-Deit-T, S and B.
We only use  DGSSA to compress the early layers of Swin-Tiny, Small and Base, since Swin has already a pyramid distribution of token numbers. The results are denoted as DGSSA-Swin-T, S and  B.
The configurations of compression are shown in Tab.~\ref{tab:configurations}. 

As \bp{shown} in Tab.~\ref{tab:benchmark}, our integrated compression pipeline can increase the run-time throughput of Deit-Tiny, Small and  Base by 73.9\%, 121.1\% and 113.0\%,  respectively. 
The increased throughput is much higher than previous state-of-the-art methods of Evo~\cite{xu2021evo}, PS~\cite{tang2021patch},  DynamicViT~\cite{rao2021dynamicvit} and TOME~\cite{bolya2022token} by a large margin. 
It is notable that Integrated-Deit-S has no accuracy drop despite making \bp{so much} acceleration. 
% Ours-Deit-B, Ours-Deit-T. 
Meanwhile, Integrated-Deit-T and B  can accelerate Deit-Small and Base much more than these methods with 0.1 \%  and 0.3\% accuracy drop. 
The comparison  verifies the effectiveness of the proposed integrated pipeline. Moreover,  new state-of-the-art compression results of Deit are  produced. 

Even though Swin has already been quite efficient,  DGSSA-based compression can   
 increase the throughput of the baselines by 9.0\%, 10.5\% and 10.0\%  with 0.0\%, 0.0\% and 0.1\% accuracy drop, respectively. 
%  accelerates Swin-Small by 10.0\% with 0.4\% accuracy drop.
% by ?, ? and  ? respectively with a little accuracy drop,
The successful compression results can verify \bp{that} the DGSSA can be applied to  
not only the global self-attention used by Deit but also
the local window attention used by Swin. 

% further accelerates Swin-Tiny/Small/Base, even though Swin has already been quite efficient. 
%  DGSSA-Swin-T, DGSSA-Swin-S and DGSSA-Swin-B 
 
% \deli{Why does Integrated outperform Evo,  DynamicViT and PS? There is no need of  predictor module of pruning token, like DynamicViT.}

% It is notable that 
% the acceleration of run-time throughput is \deli{higher than that of theoretical FLOPs} in the DGSSA-based compression of Swin. That indicates the DGSSA method is \bp{ more hardware-friendly.} 

\subsection{Ablation Studies on DGSSA}
% \subsubsection{Limited Scalability of DGSSA}
\noindent\textbf{Limited scalability of SSA.}
We make experiments to explore the scalability of DGSSA. In the experiments, DGSSA is applied to the first 2, 4, 6, 8, 10 layers of Deit-S and the first 6, 12, 18 and 20 layers of Swin-S, respectively. 

\begin{figure}[t]
  \centering
%   \fbox{\rule{0pt}{2in} \rule{0.9\linewidth}{0pt}}
    \vspace{-10pt}
    \subcaptionbox{Deit}{
  \includegraphics[width=0.47\linewidth]{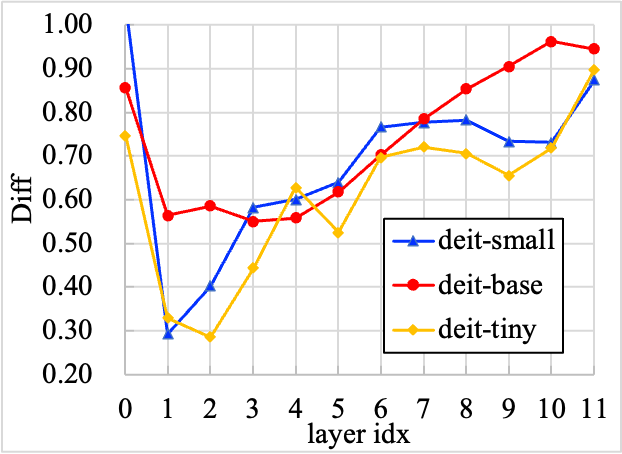}
      }
      % \hspace{-10pt}
          \subcaptionbox{Swin}{
  \includegraphics[width=0.47\linewidth]{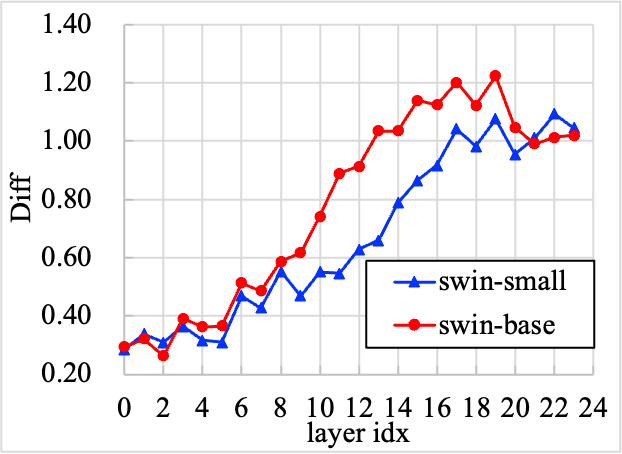}
      }
         % \vspace{-8pt}
   \caption{The metric measuring the difference between estimated static self-attention and dynamic self-attention across layers in Deit and Swin. The early layers, except 0-th layer in Deit, have smaller difference metric value than later layers in general. 
   }
   \vspace{-15pt}
   \label{fig:staic_dynamic_difference}
\end{figure}

There is no accuracy drop after DGSSA is applied in the first 2 layers of Deit-Small and in the first 6 layers of Swin-Small, as shown in  Tab.~\ref{tab:comparison_attention_1} and Tab.~\ref{tab:comparison_attention_2}, respectively.
When the number of static layers increases, the accuracy of ``DG'' decreases. When 10 layers out of total 12 layers apply DGSSA on Deit, the accuracy severely drops from 79.9\% to 78.18\% by 1.72\%, which is unbearable accuracy gap. 
The results suggest that \textit{SSA can only be applied in the few first layers}.
The conclusion accords with the difference metric in Fig.~\ref{fig:staic_dynamic_difference}, which shows that
% As shown in Fig.~\ref{fig:staic_dynamic_difference}, 
the early layers have relatively small difference metric value in general.
According \bp{to} the conclusion, 
MLP-Mixer is unreasonable because it assumes that all layers are static.

% \subsubsection{Dynamic-guided Way v.s. Others}

\begin{table}[htp]
  \begin{center}
        \caption{Dynamic-guided way outperforms other two types of static self-attention and MLPs on Deit-Small.
        ``F2Ls'' denotes the results when static self-attention or MLPs is applied in the \textbf{f}irst \textbf{2} \textbf{l}ayer\textbf{s}~(F2Ls) of Deit-Small.
        % The three types of attention are applied to the first 2, 4, 6, 8, 10 layers of Deit-S and the first 6, 12, 18 and 20 layers of Swin-S, respectively. 
        % ``MLP'', ``random'' and ``local'' refer to MLP-Mixer~\cite{tolstikhin2021mlp}, Synthesizer~\cite{tay2021synthesizer} and local positional self-attention~\cite{cordonnier2019relationship}, respectively.  
        DG refers to the proposed DGSSA. We train all models for 100 epochs for saving computation. 
        }
      \label{tab:comparison_attention_1}
      \footnotesize
  \begin{tabular}{|c|l|c|c|c|c|c|}
  \hline
  &Types & F2Ls & F4Ls & F6Ls & F8Ls & F10Ls  \\
  \hline
    \multirow{4}{*}{\makecell[c]{ Deit-S}}
  & \makecell[l]{DG}  & \textbf{79.99} & \textbf{79.73}  & \textbf{79.34} & \textbf{78.87} & \textbf{78.18}\\
%   & \makecell[l]{DG}  & 79.95 & \textbf{79.81}  & \textbf{79.00} & \textbf{78.50} & 77.8\\
  & normal & 79.96 & 79.55  & 78.99 & 78.59 & 77.90\\
  & local~\cite{cordonnier2019relationship} &79.94  &79.48 & 78.73 &77.98 &77.08\\
  & MLPs~\cite{tolstikhin2021mlp} & 79.36 & 77.90  & 75.80 & 73.84& 73.19\\
  \hline
%   \hline
%   &type & 2  & 4 & 6 & 8 & 10  \\
%     \hline
%     \multirow{3}{*}{\makecell[c]{ Deit-S-cut}} 
%   & priors & \textbf{79.91} &  &  & &\\
%   & MLPs & - & -  & - & -& -\\
%   & random & 79.50 &  &  & &\\
%   & local  & 79.40 &  &  & &\\
%   \hline
  \end{tabular}
    \vspace{-15pt}
    \end{center}
\end{table}

  \begin{table}[htp]
  \begin{center}
        \caption{Dynamic-guided way outperforms other two types of static self-attention and MLPs on Swin-Small.  {Refer to Tab.~\ref{tab:comparison_attention_1} for the meaning of denotations.} 
         We train all models for 50 epochs for saving computation. 
        % The three types of attention are applied to the first 2, 4, 6, 8, 10 layers of Deit-S and the first 6, 12, 18 and 20 layers of Swin-S, respectively. 
        % ``MLP'', ``random'' and ``local'' refer to MLP-Mixer~\cite{tolstikhin2021mlp}, Synthesizer~\cite{tay2021synthesizer} and local positional self-attention~\cite{cordonnier2019relationship}, respectively.  
        % DG refers to the proposed dynamic-guided way. 
        }
      \label{tab:comparison_attention_2}
      \footnotesize
  \begin{tabular}{|c|l|c|c|c|c|}
    \hline
  &Types & F6Ls  & F12Ls & F18Ls & F20Ls   \\
    \hline
    \multirow{4}{*}{\makecell[c]{ Swin-S}} 
  & DG & \textbf{83.00} & \textbf{82.63} & \textbf{81.68} & \textbf{81.39}\\
  & normal & 82.82 & 82.36 & 81.54 & 81.15\\
  & local~\cite{cordonnier2019relationship}&82.66  &81.93  &  80.33&  79.96\\
  & MLPs~\cite{tolstikhin2021mlp} & 82.10 & 80.64  & 77.26 & 76.51 \\
  \hline
  \end{tabular}
  \vspace{-15pt}
  \end{center}
\end{table}

\noindent\textbf{Dynamic-guided way v.s. others.} We investigate the effectiveness of the proposed DGSSA, denoted as ``DG'',  by making comparison experiments with normal, local position-aware self-attention~\cite{cordonnier2019relationship,d2021convit} and MLP-Mixer~\cite{tolstikhin2021mlp} under the comparable calculation complexity. 
 ``nomral''  trains the static self-attention matrix from random initialization. 
 Works \bp{like} \cite{cordonnier2019relationship,d2021convit}, denoted as ``local'', use a position-aware static self-attention matrix and their implementation follows the public code~\footnote{https://github.com/facebookresearch/convit}. 
 MLP-Mixer, denoted as ``MLPs'', is also a static network. 
 We make its implementation from the code~\footnote{https://github.com/rishikksh20/MLP-Mixer-pytorch}, and the token \bp{dimension} is kept default value while the channel \bp{dimension} is adjusted to ensure the comparable calculation complexity. 
 
%  Its calculation complexity is same as our method's, because the sparse attention matrices is treated as  dense matrix in matrix multiplication. 
%  because  sparse multiplication.

The results are shown in Tab.~\ref{tab:comparison_attention_1} and  \ref{tab:comparison_attention_2}, \bp{from which we can notice that} 
the proposed ``DG''  outperforms ``random'', ``MLPs'' and ``local'' under different configurations of self-attention and on two kinds of ViTs of Deit-S and Swin-S. 
% Moreover, the gains of ``DG'' are more obvious when the more layers applies static self-attention.
% The reason why dynamic-guided way outperforms ``normal'' may be that 
The useful information inherited from the replaced dynamic self-attention can serve as good 
initialization in the dynamic-guided way. By contrast, ``normal''  trains their self-attention matrices from scratch. Thus, the \textit{good initialization} may account for its effectiveness. 

Moreover,  both global ``DG'' and ``normal'' outperform ``local'' in Tab.~\ref{tab:comparison_attention_1} and  \ref{tab:comparison_attention_2}, which may result from that the local position-aware patterns may lose some global details due to the limited receptive field. 
Since the position-aware self-attention is equivalent to CNNs with kernel size of square root of head number, 
the conclusion may indicate that CNNs may not be a good choice to relieve the computation of early layers for ViTs. 
 
% directly discards it away. The difference may account for the gains. 
Finally, MLP-Mixer has the worst performance in all configurations. Then, we can make conclusion that 
making self-attention module of ViTs static is better than mixing MLP-Mixer blocks into ViTs in early layers.

\input{ablation_study_glad_version1}

\subsection{Integrated pipeline v.s. single usage}
In this section, we make comparison experiments to illustrate \bp{that} the integrated usage of DGSSA and GLAD is better than single usage. We use  Integrated-Deit-S as the baseline. Integrated-Deit-S uses DGSSA in the first two layers and \bp{applies} GLAD in the later layers~(3-th, 5-th and 7-th layer). 
{A1 and A2 are counterparts of Integrated-Deit-S under comparable throughput. } Specifically, A1 drops GLAD and still uses DGSSA in the later layers, while 
A2 drops the DGSSA and still uses GLAD in the first two layers on the base of Integrated-Deit-S.
The configurations of B1 and B2 adopt the subset of the configuration used by Integrated-Deit-S. Specifically, 
B1 drops GLAD in the later layers, while B2 drops the DGSSA in the first two layers on the base of Integrated-Deit-S.

We  \bp{can make two conclusions from} the results shown in Tab.~\ref{tab:integrated_acc}. 
Firstly,  GLAD-based compression of 
A2 \bp{causes} 0.5\% accuracy drop under comparable acceleration, while 
 DGSSA-based compression of A1 cannot produce so much acceleration.
% A2 further applies SSA in the first two layers on the base of Integrated-Deit-S, and accelerate 55\%  with no accuracy drop. 
% By contrast, A4 and A5 still use GLAD in the first two layers on the base of Integrated-Deit-S, and  
Thus, \textit{the integrated compression is much more effective than single usage under comparable throughput}.
Secondly, integrated-Deit-S further uses DGSSA in the first two layers on the base of B2 and yields {+146 img/s} without accuracy drop. Compared with B1, Integrated-Deit-S  further uses GLAD in the later layers and obtain {+1057 img/s} without accuracy drop too.
The comparison illustrates that \textit{integrated usage can make further acceleration on the base of  single usage without accuracy drop.}

% Integrated-Deit-S  uses GLAD in the 3-th, 5-th and 7-th layers, and can accelerate 50\% with a accuracy of 79.9\%. 
% A2 further applies SSA in the first two layers on the base of Integrated-Deit-S, and accelerate 55\%  with no accuracy drop. 
% By contrast, A4 and A5 still use GLAD in the first two layers on the base of Integrated-Deit-S, and cause a little accuracy drop under comparable acceleration. The similar conclusion can be made in the experiments of a series of B. Thus, 
% the integrated compression can have higher accuracy  under comparable throughput and 
% can make higher throughput with similar accuracy than single usage.

\begin{table}[ht]
  \begin{center}
    % \vspace{-8pt}
  \scriptsize
  \caption{
  Results of ablation study of integrated pipeline. Integrated-Deit-S is the baseline. ``all'' means the counterpart configurations, in which DGSSA or GLAD is applied in both early and later layers, under comparable throughput. 
  ``partial'' means configurations which are subset of configurations used by Integrated-Deit-S.  Following \cite{touvron2021training,liu2021swin}, we measure throughput  on a V100 GPU.
%   The configurations of compression are Tab.~\ref{tab:integrated_configurations}. 
  }
  \label{tab:integrated_acc}
  \begin{tabular}{|c|c|c|c|}
   \hline
      No. & Configuration & \makecell[c]{Throughput\\(img/s)} & \makecell[c]{Top-1 Acc.\\(\%)} \\
   \hline
   Integrated-Deit-S& DGSSA+GLAD & 2080  & 79.9 \\
   \hline
A1& all DGSSA & 1200 & 78.2 \\
% Integrated-Deit-S& GLAD & 50\% & 79.9\%\\
A2& all GLAD &  {2108} & 79.4 \\
% A5& GLAD & 55\%& ?\%\\
\hline
B1& partial DGSSA & 1023 & 79.9\\
B2& partial GLAD & {1934} & 79.9 \\
% B3& SSA+GLAD & & \\
% B4& GLAD & & \\
% B5& GLAD & & \\
% B6& GLAD & & \\
% B7& GLAD & & \\
\hline
  \end{tabular}
  \end{center}
  \vspace{-25pt}
\end{table}

\section{{Conclusion}}
In this paper, we observe that ViTs have heterogeneous self-attention
patterns that attention maps have more similar patterns across different images in early layers than  later layers, while have more low-rank patterns in  later layers than  early layers. 
Inspired by the observations,  we propose an integrated compression pipeline of dynamic-guided static self-attention~(DGSSA) and  global aggregation pyramid~(GLAD)  to accelerate the whole ViTs.
% method  for the acceleration of later layers. 
The advantages of  DGSSA and GLAD over previous methods are  verified by extensive ablation studies, respectively. Moreover,  
the integrated pipeline produces new SOTA results that Deit can be accelerated by up to 121\% with negligible accuracy drop. In the future, we will determine the hyper-parameters $P$ and $Q$ for the better use of the integrated pipeline by automatic machine learning techniques and other strategies. 
% in  a strategy to determine layer numbers of applying static self-attention. 

{\small
\bibliographystyle{ieee_fullname}
\bibliography{egbib}
}
\end{document}

%% file: ablation_study_glad_version1.tex
\subsection{Ablation Studies on GLAD}
In this part, we compare GLAD with other pruning-based and aggregation-based methods to illustrate the effectiveness of GLAD.

\noindent{\textbf{Comparison with pruning-based methods.}} As shown in Tab.~\ref{tab:benchmark}, GLAD-based compression outperforms other pruning-based methods, such as Evo~\cite{xu2021evo}, PS~\cite{tang2021patch},  DynamicViT~\cite{rao2021dynamicvit} and TOME~\cite{bolya2022token} in terms of throughput and accuracy on Deit-Small.
The reasons may come from two folds: 1) pruning-based methods will \bp{lose} spatial information after reducing token numbers, while GLAD can preserve the information as much as possible while reducing token spatial size by a linear aggregation. 2) GLAD is more efficient in run-time mode due to no need of  importance score prediction module which is adopted by DynamicViT~\cite{rao2021dynamicvit} during inference. 
% full number of tokens into less number by 
% The GLAD-based compression is a fair counterpart of other pruning-based methods, . GLAD-Deit-B, GLAD-Deit-S and GLAD-Deit-T produces much higher throughput and less accuracy drop than them. That verifies the effectiveness of GLAD. 

% To thoroughly validate the effectiveness of GLAD, 
\noindent{\textbf{Comparison with aggregation-based methods.}} We also make controlled comparison experiments to compare GLAD with other aggregation-based methods, such as the local concatenation used in Swin~\cite{liu2021swin} and Twins~\cite{chu2021twins}, and down-sampling used in \cite{wu2021cvt}. 
To make fair comparison, we control the same aggregated token numbers and do not use DGSSA method in the early layers in all experiments.
% We replace GLAD with these methods in Our-Deit-S, respectively. 
\begin{table}[htp]
  \begin{center}
        \caption{The comparison of GLAD with other aggregation-based methods of reducing token numbers. 
        % Concatenation is used in Swin~\cite{liu2021swin} and Twins~\cite{chu2021twins}, and  down-sampling is used in \cite{wu2021cvt}. 
        }
      \label{tab:comparison_GLAD}
      \footnotesize
  \begin{tabular}{|l|c|c|c|}
  \hline
  Methods & Scope & Top-1 Acc.(\%)   \\
  \hline
%   GLAD & global  & \textit{flexible} & 79.9 \\
  GLAD & global & 80.4 \\
  \hline
  concatenation~\cite{liu2021swin,chu2021twins} & local   & 80.0  \\
  down-sampling~\cite{wu2021cvt} & local   & 80.1  \\
  \hline
 \end{tabular}
 \end{center}
  \vspace{-15pt}
 \end{table}
 
%  \begin{table}[htp]
%   \begin{center}
%         \caption{The comparison of GLAD with other aggregation-based methods of reducing token spatial size. 
%         Concatenation is used in Swin~\cite{liu2021swin} and Twins~\cite{chu2021twins}, and  down-sampling is used in \cite{wu2021cvt}. }
%       \label{tab:comparison_GLAD}
%       \footnotesize
%   \begin{tabular}{|l|c|c|c|}
%   \hline
%   Methods & Scope & Ratio & Top-1 Acc.(\%)   \\
%   \hline
% %   GLAD & global  & \textit{flexible} & 79.9 \\
%   GLAD & global  & \textit{fixed} at $\frac{1}{4}$ & 80.4 \\
%   \hline
%   concatenation & local& \textit{fixed} at $\frac{1}{4}$   & 80.0  \\
%   down-sampling & local& \textit{fixed} at  $\frac{1}{4}$   & - \\
%   \hline
%  \end{tabular}
%  \end{center}
%   \vspace{-15pt}
%  \end{table}
 
 As shown in Tab.~\ref{tab:comparison_GLAD}, GLAD outperforms concatenation and down-sampling methods.
%  Two features of ``global'' and ``flexible'' may account for the performance.
%  1)
%  That may result from that 
Global GLAD has improvement of {0.4 \% and 0.3 \% Top-1 accuracy} over local concatenation and down-sampling methods 
% with fixed $\frac{1}{4}$ reduction ratio of token number, 
, respectively.
The feature of ``global'' may account for the performance. 
% That indicates the necessity of global scope. 
% 2) Moreover, the improvement further comes to  0.x \% and 0.x \%   when GLAD flexibly \bp{uses} proper compression configuration of slimmed token number. 
% That indicates the effectiveness of flexible aggregation.
The two local methods aggregate tokens to get less token numbers  within a fixed-size window, such as $2\times 2$.  
The pre-defined fixed size may not 
be suitable for the real distribution of feature redundancy information. 
Above all, the comparison results validate the \textit{effectiveness of GLAD which   makes global aggregation.}